%% file: wacv.tex
\documentclass[10pt,twocolumn,letterpaper]{article}

\usepackage{wacv}
\usepackage{times}
\usepackage{epsfig}
\usepackage{graphicx}
\usepackage{amsmath}
\usepackage{amssymb}

\usepackage[dvipsnames]{xcolor}
\usepackage[numbers,sort,compress]{natbib}  % sort the reference
\makeatletter
\@namedef{ver@everyshi.sty}{}
\makeatother
\usepackage{tikz}

\usepackage{comment}
\usepackage{color}
\usepackage{booktabs}
\usepackage{multirow}
\usepackage{pifont}
\usepackage{listings}

\usepackage{makecell}
\usepackage{fancyhdr}
\usepackage[ruled,vlined]{algorithm2e}
\usepackage{tabularx}
\newcolumntype{Y}{>{\centering\arraybackslash}X}

\makeatletter
\DeclareRobustCommand\onedot{\futurelet\@let@token\@onedot}
\def\@onedot{\ifx\@let@token.\else.\null\fi\xspace}

\def\eg{\emph{e.g}\onedot} 
\def\ie{\emph{i.e}\onedot}

\makeatother

\usepackage[accsupp]{axessibility}  % Improves PDF readability for those with disabilities.

\def\wacvPaperID{154} % Enter the WACV Paper ID here

\wacvalgorithmstrack   % Uncomment this line if you are submitting to the Algorithms Track.
\wacvfinalcopy % *** Uncomment this line for the final submission

\ifwacvfinal
\usepackage[breaklinks=true,bookmarks=false]{hyperref}
\else
\usepackage[pagebackref=true,breaklinks=true,colorlinks,bookmarks=false]{hyperref}
\fi

\begin{document}

%%%%%%%%% TITLE
\title{Self-Supervised Pyramid Representation Learning
\\for Multi-Label Visual Analysis and Beyond}

\author{Cheng-Yen Hsieh\\
National Taiwan University\\
% No. 1, Sec. 4, Roosevelt Rd., Taipei, Taiwan\\
{\tt\small chengyeh@andrew.cmu.edu}
\and
Chih-Jung Chang\\
National Taiwan University\\
{\tt\small b06201018@ntu.edu.tw}
\and
Fu-En Yang\\
National Taiwan University\\
{\tt\small f07942077@ntu.edu.tw}
\and
Yu-Chiang Frank Wang\\
National Taiwan University\\
{\tt\small ycwang@ntu.edu.tw}
}

\maketitle
\thispagestyle{empty}

%******************* Abstract
\input{0_abstract.tex}

%%%%%%%%% BODY TEXT
\input{1_introduction.tex}

%------------------------------------------------------------------------
\input{2_related.tex}

%------------------------------------------------------------------------
\input{3_method.tex}
\input{pseudocode.tex}
\input{4_experiment.tex}

%------------------------------------------------------------------------
\input{5_conclusion.tex}

{\small
\bibliographystyle{ieee_fullname}
\bibliography{egbib}
}

\end{document}

%% file: 0_abstract.tex
\begin{abstract}

% \textcolor{blue}{Most existing self-supervised pretext tasks are designed for single-label image classification tasks by describing an entire image as a single vector. However, such methods might not be sufficient to observe the presence of multiple objects or to model the inherent semantic structure in an image, which limit the transfer performance on multi-label visual analysis tasks.}
While self-supervised learning has been shown to benefit a number of vision tasks, existing techniques mainly focus on image-level manipulation, which may not generalize well to downstream tasks at patch or pixel levels. Moreover, existing SSL methods might not sufficiently describe and associate the above representations within and across image scales. In this paper, we propose a Self-Supervised Pyramid Representation Learning (SS-PRL) framework. The proposed SS-PRL is designed to derive pyramid representations at patch levels via learning proper prototypes, with additional learners to observe and relate inherent semantic information within an image. In particular, we present a cross-scale patch-level correlation learning in SS-PRL, which allows the model to aggregate and associate information learned across patch scales. We show that, with our proposed SS-PRL for model pre-training, one can easily adapt and fine-tune the models for a variety of applications including multi-label classification, object detection, and instance segmentation.

% \keywords{Self-Supervised Learning, Model Pre-training, Multi-Label Visual Analysis}
\end{abstract}

%% file: 1_introduction.tex
\section{Introduction}

\label{sec:intro}

% 1-1
To understand the complex relations in natural scenes or explore rich information from an image, many real-world visual recognition tasks (\textit{e.g.}, semantic scene classification~\cite{wang2008automatic}, or medical diagnosis~\cite{abbas2013pattern}) require the learned model to predict \emph{more than one} semantic label given a single input image. The conventional single-label classification methods mainly focus on assigning \emph{single} class label to each image without considering the multiple-object scenarios in one image or handling the relations among distinct label semantics. More particularly, the derived features are required to describe the presence of multiple objects and semantic label dependencies in an image for tackling multi-label visual analysis tasks. While existing~\cite{zhang2010multi,yeh2017learning,wang2016cnn,wu2020distribution,liu2021query2label,ridnik2021imagenet,ridnik2021asymmetric,ridnik2021ml} methods perform promising performance, they still acquire a large amount of multi-label annotated data for training. Considering the labeling cost, collecting fully annotated data for learning a model for multi-label tasks would be computationally expensive.    

% Multi-Label classification \cite{gibaja2015tutorial}, where each instance is assigned with multiple labels, is a challenging problem in computer vision with a variety of applications ranging from medical diagnosis, video annotation, and semantic scene classification\cite{abbas2013pattern, 1677522,wang2008automatic}. Different from standard multi-class
% classification problems (i.e., only one single label for each
% input image), multi-label classification typically requires exploring the correlation between different labels to produce satisfactory performances \cite{zhang2010multi, yeh2017learning, wang2016cnn}. Many studies have also been conducted to deal with the non-uniformity of spatial scales between the objects in each image by utilizing multi-scale input images \cite{durand2017wildcat, wang2016beyond}. Unfortunately, the outstanding performance in these methods typically relies on the availability of a large amount of labeled data, which is time-consuming and not practically feasible to acquire. 
        
    %Modern development of deep neural network (DNN) has achieved great success on various challenging computer-vision tasks, such as image recognition \cite{ILSVRC15}, multi-label classification \cite{tsoumakas2009mining}, and object detection \cite{lin2014microsoft}. Unfortunately, The outstanding performance in these tasks typically relies on the availability of a large amount of labeled data, which is time-consuming and not practically feasible to acquire. %

% 1-2
To alleviate the huge burdens of collecting and annotating large-scale multi-label datasets, an effective approach is to pre-train a general-purpose model in the self-supervised learning (SSL) manner, followed by the fine-tuning process to facilitate the learning of downstream tasks of interest.   
Recent SSL pre-training approaches~\cite{chen2020simple,he2020momentum,chen2020improved,grill2020bootstrap,caron2020unsupervised,mitrovic2020representation,dwibedi2021little,lee2021compressive,bachman2019learning,van2021revisiting} learn discriminative representations based on \emph{image}-level contrastive learning scheme, which pulls the views from the same image together and pushes the features from different images away. While such training fashion significantly improves the performance on single-label image classification, the above SSL methods are only trained at \emph{image}-level, which lacks the ability to describe the multiple objects in an image.
Hence, transferring the learned knowledge from such SSL pre-trained models to downstream multi-label visual analysis tasks remains underexplored.

% To relieve the demands of data annotations, extensive studies have been dedicated to learning from unlabeled data. Within these efforts, self-supervised learning (SSL) of visual representation becomes a well-established pretraining technique \cite{Jaiswal_2020} facilitating efficient training on downstream tasks. Most SSL methods design different pretext tasks to utilize unlabeled data and learn representations that are invariant under different data augmentations. Many methods \cite{he2020momentum, chen2020improved, grill2020bootstrap, caron2020unsupervised} have achieved comparable results compared to the fully-supervised method for downstream image classification tasks. Nevertheless, most of these works do not take into account the complex relationship in the label space and the variability of spatial scales which are both crucial in multi-label classification tasks.

% 1-3
To perform pre-training for downstream multi-label tasks, we aim at exploiting inherent semantic label dependencies in a \textit{self-supervised} manner. In this paper, we propose a unique \emph{self-supervised pyramid representation learning (SS-PRL)} framework. Without observing any ground truth labels at either image or object levels, our SS-PRL is learned in a \emph{cross-scale patch-level SSL} manner that derives pyramid representations and semantic prototypes at \emph{patch} levels. This allows one to explore the presence of objects and label dependencies in an image while leveraging the correlation across multiple patch scales to associate and aggregate the knowledge learned from different patch scales.
% Without observing any ground truth class image or object-level labels, our SS-PRL jointly learns pyramid representations across multiple scales and derives semantic prototypes, which would observe and preserve intrinsic semantic label structures from training image data. 

With the particular aim of exploiting fine-grained information within an image for mimicking objects presented at various scales, our proposed SS-PRL constructs multiple branches to extract global image-level and local patch-level features from the input image for learning the pyramid representations and associated prototypes. These prototypes are designed to serve as semantic cues for describing label dependencies and thus are expected to improve the model capability for downstream multi-label tasks (\textit{e.g.,} multi-label image classification, or object detection). 

To further integrate the information from different patch-level representations, we present~\emph{cross-scale patch-level correlation learning} in SS-PRL. This enforces the correspondence of output predictions from global image and local patches, which guides the model to leverage multi-grained information. To verify the effectiveness of our SS-PRL for diverse downstream tasks, we consider multi-label image classification, object detection, and segmentation benchmarks in our experiments. We confirm that our SS-PRL performs favorably against SOTA methods and achieve promising performances.

The contributions of our work are highlighted below:
\begin{itemize}
    \item To the best of our knowledge, we are among the first to design pretext tasks in a self-supervised manner for facilitating downstream multi-label visual analysis tasks. 
        
    \item We propose Self-Supervised Pyramid Representation Learning (SS-PRL), deriving multi-scale patch-level pyramid representations with semantic prototypes discovered to exploit their inherent correlation.
        
    \item A unique cross-scale patch-level correlation is introduced in our SS-PRL to leverage the learned knowledge across multiple and distinct spatial scales, ensuring sufficient representation ability of our model.
        
    \item In addition to a wide range of downstream tasks at object instance and pixel levels, we qualitatively demonstrate that the learned prototypes at different scales would describe the associated visual concepts.  
        
        % Particularly, we achieve \textcolor{red}{XX\%} and \textcolor{red}{XX\% mAP} on COCO~\cite{lin2014microsoft} and Pascal VOC~\cite{everingham2010pascal} with a standard ResNet.

\end{itemize}

% \begin{itemize}
%     \item A brief introduction of self-supervised learning and its success in image classification task.
%     \item Recent researches start focusing on different downstream tasks.
%     \begin{itemize}
%         \item DenseCL (CVPR2021): Dense prediction task.
%         \item SCRL (CVPR2021): Localization tasks such as object detection and instance segmentation.
%         \item DetCo (ICCV2021): Object detection.
%     \end{itemize}
%     \item However, there is no previous work designed for multi-label classification. (ref DetCo: There is no single best pretext task for different downstream tasks.)
%     \item Introduce the proposed framework ML-SSL.
%     \item Contribution
%     \begin{itemize}
%         \item 3 strategies for multi-label-friendly pretext task
%         \item Propose ML-SSL
%         \begin{itemize}
%             \item Local clustering
%             \item Local-to-global loss
%         \end{itemize}
%         \item Results on the downstream task.
%     \end{itemize}
% \end{itemize}

%% file: 2_related.tex
\section{Related work}

%\paragraph{Multi-Label Image Classification.}
\subsection{Multi-Label Image Classification}
Multi-label image classification aims at assigning a set of labels to each image. Due to the fact that pictures in everyday life are inherently multi-labeled and contain more complex visual appearances and diverse label semantics, multi-label visual analysis is more practical yet challenging compared with conventional single-label classification tasks. Associating local image regions to labels has been proven to be beneficial in multi-label classification since an image is usually composed of objects with different scales located in arbitrary regions. 
SRN~\cite{zhu2017learning} learns an attention map that associates related image regions to each label in order to portray the underlying spatial relation between semantic labels. Gau \textit{et al.}~\cite{guo2019visual} improves the performance of multi-label classification by introducing a consistency objective on visual attention regions under image transformations. 
In addition, Ridnik \textit{et al.}~\cite{ridnik2021asymmetric} and Wu \textit{et al.}~\cite{wu2020distribution} propose asymmetric loss and distribution-balanced loss respectively to mitigate the accuracy degradation from the positive-negative imbalance.
While promising, most existing works~\cite{zhu2017learning,guo2019visual,ridnik2021asymmetric,wu2020distribution,liu2021query2label,ridnik2021imagenet,ridnik2021ml} generally learn the correspondence between image regions and labels in a fully-supervised fashion.
% In contrast of the fully-supervised multi-label image classification approaches, our method aims to learn in a \emph{self-supervised} fashion that associates local image regions with pseudo labels assigned by our multi-level semantic prototypes.

To mitigate the costly process of collecting and annotating large-scale multi-labeled datasets, various settings of multi-label classification with limited supervision have been proposed. For example, multi-label learning with missing labels~\cite{sun2010multi} considers the case in which only a partial set of labels is available; semi-supervised multi-label classification \cite{chen2008semi} admits a few fully-labeled data and a large amount of unlabeled data; partial multi-label learning \cite{xu2020partial} discusses the setting that each instance is annotated with a set of candidate labels. 
Different from the above settings, we aim to tackle multi-label visual analysis in a \emph{self-supervised} fashion that pre-trains on unlabeled data while only using a few labeled samples for further fine-tuning. 
% associates local image regions with pseudo labels assigned by our multi-level semantic prototypes for downstream multi-label classification tasks.
% ------------------------------------------------
%\paragraph{Self-Supervised Learning.}

\begin{figure*}[t]
\begin{center}
\includegraphics[width=0.85\linewidth]{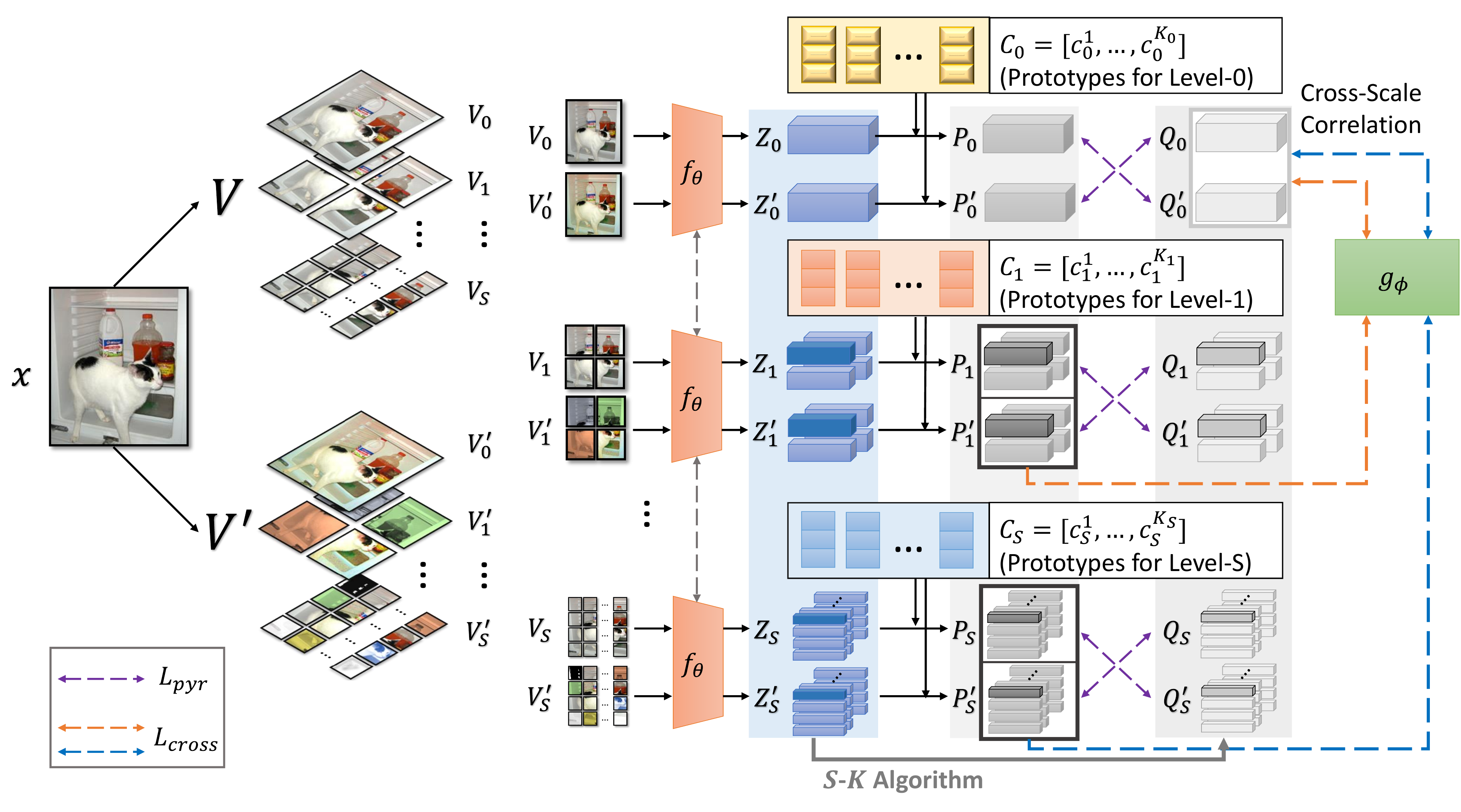}
\end{center}
\vspace{-3mm}
\caption{\textbf{Self-Supervised Pyramid Representation Learning.} The input $x$ is augmented into two pyramid views $V=\{V_{s}\}_{s=0}^{S}$ and $V^{\prime}=\{V_{s}^{\prime}\}_{s=0}^{S}$ with patch sets obtained at each scale. For scale $s$, we have $f_{\theta}$ derive the pyramid representations $Z$, which are further transformed into prototype-based representations $P$ based on the learned/assigned prototypes $C_s$ at that scale. With the prototype assignments $Q$ inferred from $Z$ via S-K algorithm~\cite{cuturi2013sinkhorn}, we observe correlation between $Q$ and the aggregated $P$ across each scale via $g_{\phi}$ as cross-scale patch-level self-supervision.}
\vspace{-3mm}
\label{fig:framework}
\end{figure*}

\subsection{Self-Supervised Learning}\label{ssec:ssl} 
Recently, self-supervised learning methods~\cite{bachman2019learning,misra2020self,chen2020simple,he2020momentum,chen2020improved,grill2020bootstrap,dwibedi2021little,lee2021compressive,van2021revisiting,mitrovic2020representation,chen2021exploring,zbontar2021barlow,asano2020self,caron2020unsupervised,bardes2021vicreg,caron2021emerging,li2020prototypical,goyal2021self,van2020scan} achieve remarkable progress on single-label image classification and narrow the performance gap compared with fully-supervised counterparts. One group of SSL approaches adopt the contrastive objective to perform instance discrimination on a large amount of unlabeled data. For instance, PIRL~\cite{misra2020self}, SimCLR~\cite{chen2020simple}, and MoCo v1/v2~\cite{he2020momentum,chen2020improved} share the same concept of pulling multiple views of an image close while pushing different instances apart to derive the compact yet discriminative representations. BYOL~\cite{grill2020bootstrap} and SimSiam~\cite{chen2021exploring} claim that the use of asymmetry network architecture and exponential moving average update strategy are the crucial factors in preventing mode collapsing when the training process only rely on positive pairs. Barlow Twins~\cite{zbontar2021barlow} tries to align the corresponding entities between the two embedded features from each positive pair by Siamese network.
% \textcolor{red}{However, the methods mentioned above only enforce the representations of augmented views of an image to be similar, while not defining the behavior of, or even pushing away, the representations of different images, which could share certain semantic meanings. As a result, they tend to ignore the hidden structure of a dataset.}

Another group of SSL works can be viewed as clustering-based methods, which learn visual representations via pseudo-label prediction. DeepCluster~\cite{caron2018deep} and SeLa~\cite{asano2020self} apply k-means clustering and optimal transport respectively to produce pseudo labels. In contrast to \cite{caron2018deep,asano2020self}, SwAV~\cite{caron2020unsupervised} proposes an online clustering method that assigns the soft labels to the input image via the learned prototype vectors.
We note that the aforementioned SSL methods simply extract a single feature to represent an image and thus do not handle the presence of multiple objects from an image well. The ability to transfer the learned knowledge from such pre-training tasks (\textit{i.e.,} image-level contrastive learning or clustering) to downstream tasks with multiple labels (\textit{e.g.,} semantic segmentation, object detection, or multi-label classification) remains challenging and still underexplored.
% Although such clustering-based methods can capture the label-like information inherited in a dataset, they still hard to directly transfer the learned knowledge to multi-label classification as a downstream task since they do not take objects at different spatial scales in an image into consideration.

To better finetune the pre-training models for facilitating downstream visual classification tasks, a number of works~\cite{wang2021dense,roh2021spatially,mishra2021object,henaff2021efficient,xie2021detco,zhao2021self,yang2021instance,xie2021propagate,xiao2021region,o2020unsupervised} design specific pretext tasks which are consistent with the characteristics of downstream tasks of interest. These methods are generally dedicated to constructing the pretext tasks that benefit dense prediction like semantic segmentation, object detection, or keypoints detection. For example, DenseCL~\cite{wang2021dense} introduces a pairwise contrastive loss at the pixel-level features between two views of an input image. DetCo~\cite{xie2021detco} jointly learns discriminative representations from global images and local patches via contrastive learning across multiple scales and network layers. InsLoc~\cite{yang2021instance} proposes a localization pretext task with the contrastive loss by taking crops of foreground images pasted onto different background images. MaskCo~\cite{zhao2021self} contrasts region-level features with the contrastive mask prediction task.

We note that, while~\cite{wang2021dense,roh2021spatially,mishra2021object,henaff2021efficient,xie2021detco,zhao2021self,yang2021instance,xie2021propagate,o2020unsupervised} integrate the local information into instance discrimination scheme, they are \textit{not} designed to observe the inherent relations among objects and thus are sub-optimal for downstream multi-label visual analysis tasks.
In this paper, we design the pretext task for multi-label image classification by deriving the pyramid representations across multiple scales, producing multi-level semantic prototypes to exploit the label relations from the observed training data.   

% While the attention to local regions is addressed by these methods due to the enforced local representation consistency, the underlying relationship among objects in images is overlooked. Our work aims to design an unsupervised pretext task for multi-label classification that is aware of both objects at different scales and their label relationship.

%% file: 3_method.tex
\section{Proposed Method}
\label{sec:method}

\subsection{Problem Formulation}

For the sake of completeness, we first define the problem setting considered in this work. Given an unlabeled dataset $\mathcal{D}_u=\{x_1,x_2,...,x_N\}$ of $N$ images, we aim to learn a feature extractor $f_\theta$ on $\mathcal{D}_u$, facilitating downstream tasks associated with multi-labels. As depicted in Fig.~\ref{fig:framework}, we present a Self-Supervised Pyramid Representation Learning (SS-PRL) framework, which consists of a feature extractor $f_\theta$ and a cross-scale correlation learner $g_\phi = \{g_{\phi,s}\}_{s=1}^{S}$ at each patch scale $s$. 
We apply $f_\theta$ to derive pyramid representations $Z$ from the pyramid of views $V$, which are then transformed into prototype-based representations $P$ based on semantic prototypes $C_s$ learned at each scale. To further leverage multi-grained information from different scales, cross-scale patch-level correlation is enforced between S-K based prototype assignments $Q$ and the aggregated $P$ across scales via $g_\phi$.
Once the learning is complete, one can apply and fine-tune $f_\theta$ for downstream tasks like multi-label image classification, object detection, or segmentation.

%------------------------------------------------------------------------
\subsection{Self-Supervised Pyramid Representation Learning}

As illustrated in Fig.~\ref{fig:framework}, the framework of our proposed Self-Supervised Pyramid Representation Learning contains the stage of \emph{patch-based pyramid representation learning} and 
\emph{cross-scale patch-level correlation learning}. The former is to derive pyramid representations via prototypes learned at each \emph{patch} level, aiming to handle the presence of multiple objects while exploring the label dependencies from unlabeled data. As for the latter stage, we further associate and aggregate the learned knowledge across different patch scales by enforcing the coherence between the prediction of local patches and global images. We now detail the designs for the above two stages below. \\

\noindent \textbf{3.2.1\quad Learning of patch-based pyramid representation\\} \label{sssec:pyramid}
As noted in Section~\ref{ssec:ssl}, prior SSL works~\cite{misra2020self,chen2020simple,he2020momentum,asano2020self,caron2020unsupervised} generally embed an image into a single feature and are not designed for observing multiple objects presented in an image. Hence, such derived models and representations cannot be easily transferred to downstream multi-label visual analysis tasks. To handle images with multi-objects/labels, we derive pyramid representations at the patch level instead of producing image-level features. This allows the model to observe and capture more fine-grained information from an image. In addition, our SS-PRL is designed to learn prototypes at each patch level, which exploits potential label dependencies in an unsupervised fashion.

As shown in Figure~\ref{fig:framework}, we first build two pyramids of views $V=\{V_{s}\}_{s=0}^{S}$ and $V^{\prime}=\{V_{s}^{\prime}\}_{s=0}^{S}$, which are generated with different augmentations from the input image $x$. 
% \textcolor{blue}{To be more specific, given an input image $x$, it is transformed to two pyramids of views $\{V_{s}\}_{s=0}^{S}$ and $\{V_{s}^{\prime}\}_{s=0}^{S}$ that both consist of $S + 1$ groups of image patches corresponding to different spatial scales.}
For each patch scale $s$, the image patch group $V_s = [v_{s,1},\dotsc,v_{s,M_s}]$ is produced by dividing the image $x$ into $M_s$ non-overlapping patches and randomly transforming each patch with data augmentations.
% In particular, we set $M_0 = 1$ to represent the augmented global view \textcolor{blue}{$V_0 = [v_{0,1}]$} of the image $x$, and then monotonically increase $M_s$ to obtain patches of smaller spatial scales \textcolor{blue}{for the patch group $V_s$ at each level}.
Similar remarks can be applied to the derivation of $\{V_{s}^{\prime}\}_{s=0}^{S}$.
To derive the patch-level pyramid representations $Z_s=[z_{s,1},\dotsc,z_{s,M_s}]\in\mathbb{R}^{D\times{M_s}}$  and $Z_s^{\prime}=[z_{s,1}^{\prime},\dotsc,z_{s,M_s}^{\prime}]\in\mathbb{R}^{D\times{M_s}}$, we feed the two pyramid views $\{V_{s}\}_{s=0}^{S}$ and $\{V_{s}^{\prime}\}_{s=0}^{S}$ into feature extractor $f_\theta$, which contains a shared backbone network and $S+1$ independent projection heads corresponding to patch scale $s=0,1,...,S$. \\

% \textcolor{blue}{The feature extractor $f_\theta$ contains a shared backbone network (\ie ResNet-50) and $S+1$ independent projection heads for level $s=0,1,...,S$.} The two feature representation \textcolor{blue}{matrices $Z_s=[z_{s,1},\dotsc,z_{s,M_s}]$ and $Z_s^{\prime}=[z_{s,1}^{\prime},\dotsc,z_{s,M_s}^{\prime}]$} are extracted from \textcolor{blue}{the patches in $V_s$ and $V_s^{\prime}$} by the \textcolor{blue}{backbone network and the corresponding projection head for level $s$.} \textcolor{blue}{Note that $Z_s\in\mathbb{R}^{D\times{M_s}}$ since we embed all the extracted representations into $D-$dimensional space regardless of the input patch sizes.}

\noindent\textbf{Prototype-based self-supervised learning.}
With pyramid representations $Z_s$ and $Z_s^{\prime}$ obtained, we require our SS-PRL to produce representations that are discriminative and be capable of capturing the inherent semantic dependencies observed from training data, which is thus beneficial to downstream multi-labeled tasks.
Inspired by~\cite{caron2020unsupervised}, we learn a group of patch-level semantic prototypes $C_{s}\in \mathbb{R}^{D \times K_s}$ at each patch scale $s$ (where $K_s$ denotes the number of prototypes at scale $s$) to mine and reflect the label semantics observed from unlabeled training data. 
To allow the feature extractor $f_\theta$ and semantic prototypes $C_s$ to be learned jointly in an online fashion, we utilize the consistency between the probability distribution of $z_{s,m}$ and $z_{s,m}^{\prime}$ as self-supervision~\cite{caron2020unsupervised}. 
To be more specific, such prototypes $C_{s}$ can be viewed as clustering centroids, and we then transform $z_{s, m}$ to prototype-based representations $P_s$ by assigning each representation $z_{s, m}$ to prototypes $C_{s} = [ c_{s}^1,\dotsc,c_{s}^{{K_s}} ]$ at each scale $s$. 
We derive the prototype-based representations $P_s= [p_{s,1},\dotsc,p_{s,M_s}]\in \mathbb{R}^{K_s \times M_s}$ from $z_{s, m}$ and $C_{s}$ to represent probability distribution as follows:
\begin{equation}
\label{eqn:probability}
p_{s,m}^{\top} = \mathit{softmax}(\frac{1}{\tau}z_{s,m}^{\top}C_s),
\end{equation}
where $\tau$ is a temperature parameter as noted in~\cite{wu2018unsupervised}. 

However, simply aligning the prototype-based representations $P_{s}$ and $P^{\prime}_{s}$ might lead to mode collapse problems~\cite{caron2020unsupervised}. To alleviate this issue, we further utilize the iterative Sinkhorn-Knopp algorithm \cite{cuturi2013sinkhorn}, denoted by $S\text{-}K(\cdot,\cdot)$, to compute the prototype assignment vector $q_{s, m} = S\text{-}K(z_{s, m}, C_s)$ for two S-K based prototype assignments $Q_s = [q_{s,1},\dotsc,q_{s,M_s}]$ and $Q_s^{\prime} = [q_{s,1}^{\prime},\dotsc,q_{s,M_s}^{\prime}]$, which serve as the target of prediction by $P_s$. With the equal partition property imposed by Sinkhorn-Knopp algorithm, the consistency enforced between $p_{s,m}$ and $q_{s,m}^{\prime}$ is capable of alleviating the mode collapse problems~\cite{caron2020unsupervised}. As a result, the objective for our pyramid representation learning $L_{pyr}$ is defined as:
\begin{equation}
\label{eqn:L_pyr}
% \textcolor{blue}{L_{pyr}(V, V^{\prime}) = \sum_{s=0}^S{\sum_{m=1}^{M_s} \frac{\alpha_s}{M_s} {L_{cls}(z_{s,m}, z_{s,m}^{\prime})}},}
% \textcolor{blue}{L_{pyr} = \sum_{s=0}^S{\alpha_sL_{cls}(Z_{s}, Z_{s}^{\prime})},}
L_{pyr} = \sum_{s=0}^S \sum_{m=1}^{M_s} \frac{\alpha_s}{M_s}( \mathit{CE}(q_{s,m}^{\prime}, p_{s,m}) + \mathit{CE}(q_{s,m}, p_{s,m}^{\prime})),
\end{equation}
where $\mathit{CE}$ denotes the cross-entropy loss, and $\alpha_s$ balances each loss term at different patch scales $s$. 

% It is worth noting that, utilizing multi-level semantic prototypes ensures the feature extractor $f_\theta$ to learn regional or even fine-grained semantics across various scales from an image, which would be beneficial for downstream tasks associated with multi-labeled images.

% \textcolor{blue}{Utilizing patch-level semantic prototypes ensures the feature extractor $f_\theta$ to learn regional or potentially fine-grained semantics, but the correlation between information extracted from different patch scales could be underexplored due to the independence between each scale. As a result, we further address this issue with \emph{cross-scale patch-level correlation learning}in the following section.}

Although the above pyramid representations can be learned without label supervision, self-supervision at each scale is observed separately. As later verified in Table~\ref{table:ablation}, this would lack the ability to associate patch-level prototypes across image scales and thus limit the downstream classification tasks associated with multi-labels. This is why the additional self-supervision across patch scales needs to be enforced, as we introduce below. \\

%if we only apply prototype-based self-supervised learning at \emph{each} patch level independently, the knowledge derived from different patch levels might be mismatched and possibly lead to performance degradation. Thus, the \emph{cross-scale patch-level correlation learning} is introduced and detailed in the following section.}

%------------------------------------------------------------------------
\begin{table*}[t]
\begin{center}
\scalebox{0.70}{
\begin{tabularx}{1.38\textwidth}{ll *4{Y}}
\toprule
 &  & \multicolumn{4}{c}{Multi-Label Classification (mAP)} \\ 
\cmidrule(l){3-6}
 &  & \multicolumn{2}{c}{Pretrained on COCO} & \multicolumn{2}{c}{Pretrained on ImageNet} \\
 \cmidrule(lr){3-4} \cmidrule(l){5-6}
Pre-training Method &  & COCO & VOC & COCO & VOC \\
\midrule
Supervised &  & 62.5 & 81.8 &  68.5 & 86.7 \\ 
\midrule
MoCo v2 \cite{he2020momentum} & \multirow{3}{*}{\shortstack{\textit{general-purpose}\\\textit{SSL}}} & 50.2 & 67.9 & 54.3 & 82.5 \\
SwAV \cite{caron2020unsupervised} & & 60.3 & 79.2 & 60.1 & 83.2 \\
BYOL \cite{grill2020bootstrap} & & 52.6 & 70.1 & 58.4 & 80.2 \\
\midrule
DenseCL \cite{wang2021dense} & \multirow{4}{*}{\shortstack{\textit{dense prediction}\\\textit{SSL}}} & 57.0 & 75.2 & 60.5 & 82.9 \\
DetCo \cite{xie2021detco} &  & 52.7 & 70.6 & 60.0 & 81.3 \\ 
MaskCo \cite{zhao2021self} & & 51.9 & 70.2 & 50.3 & 75.1 \\ 
InsLoc\cite{yang2021instance} & & 45.0 & 61.8 & 49.5 & 74.8 \\ 
\midrule
\textbf{\textbf{SS-PRL (ours)}} & & \textbf{61.3} & \textbf{80.5} & \textbf{63.8} & \textbf{85.4} \\
\bottomrule
\end{tabularx}}
\end{center}
\caption{\textbf{Performance on multi-label classification tasks with fine-tuned linear classifiers on VOC and COCO.} With the backbone network (\ie ResNet-50) pre-trained with different supervised/self-supervised methods, we report the mAP on COCO and VOC with \textit{fine-tuned linear classifiers}. All methods are pre-trained on COCO with 200 epochs or ImageNet with 100 epochs, respectively.}
\vspace{-0mm}
\label{table:main}
% \vspace{-2mm}
\end{table*}

\begin{table*}[t]
\begin{center}
\scalebox{0.70}{
\begin{tabularx}{1.38\textwidth}{ll *6{Y}}
\toprule
 & & \multicolumn{6}{c}{Multi-Label Classification on COCO (mAP)} \\ 
\cmidrule(l){3-8}
 & & \multicolumn{3}{c}{Pretrained on COCO} & \multicolumn{3}{c}{Pretrained on ImageNet} \\
\cmidrule(lr){3-5} \cmidrule(l){6-8}
Pre-training Method &  & \shortstack{1\%\\labels} & \shortstack{10\%\\labels} & \shortstack{100\%\\labels} & \shortstack{1\%\\labels} & \shortstack{10\%\\labels} & \shortstack{100\%\\labels} \\
\midrule
Random Init. & & 4.6 & 10.7 & 42.5 & 4.6 & 10.7 & 42.5 \\ 
\midrule
MoCo v2 \cite{he2020momentum} & \multirow{3}{*}{\shortstack{\textit{general-purpose}\\\textit{SSL}}} & 34.0 & 46.9 & 54.2 & 26.4 & 55.8 & 63.7 \\
SwAV \cite{caron2020unsupervised} & & 43.6 & 56.2 & 61.4 & 39.3 & 58.6 & 66.9 \\
BYOL \cite{grill2020bootstrap} & & 35.1 & 48.0 & 54.8 & 38.7 & 53.1 & 62.5 \\
\midrule
DenseCL \cite{wang2021dense} & \multirow{4}{*}{\shortstack{\textit{dense prediction}\\\textit{SSL}}} & 42.9 & 54.8 & 62.2 & \textbf{43.4} & 59.4 & 65.8 \\
DetCo \cite{xie2021detco} & & 32.0 & 48.3 & 54.7 & 37.9 & 56.2 & 62.7 \\ 
MaskCo \cite{zhao2021self} & & 31.6 & 48.0 & 57.4 & 24.0 & 53.2 & 62.1 \\ 
InsLoc\cite{yang2021instance} & & 29.0 & 43.9 & 53.5 & 36.1 & 56.6 & 66.5 \\ 
\midrule
\textbf{SS-PRL (ours)} & & \textbf{45.1} & \textbf{57.0} & \textbf{62.9} & 41.0 & \textbf{60.9} & \textbf{67.4} \\
\bottomrule
\end{tabularx}}
\end{center}
\caption{\textbf{Performance on multi-label classification tasks in semi-supervised settings on COCO.} Methods listed are pre-trained on COCO for 200 epochs or ImageNet for 100 epochs, respectively. Models are then fine-tuned on 1\%, 10\%, and 100\% of labeled data randomly chosen from COCO for 20 epochs. Note that, \textit{Random Init.} denotes the model trained from scratch.}
\label{table:semi}
\vspace{-4mm}
\end{table*}

\noindent \textbf{3.2.2\quad Cross-scale patch-level correlation learning\\} \label{sssec:joint}
% \subsubsection{Cross-Level Consistency Learning} \label{sssec:joint}
As noted above, it would be desirable to train deep learning models which exploit semantic dependencies not only at each patch scale but also discover such properties with information properly aggregated and leveraged across scales. To achieve this goal and to benefit downstream multi-label classification tasks, we uniquely observe the correlation observed between the prototype/cluster assignments derived at the coarsest image scale (\ie, $Q_0$ or $Q_0^{\prime}$) and the prototype-based representations $P_s$ aggregated at each scale $s$. With a deployed cross-scale correlation learner $g_\phi$, the above correlation can be enforced and be served as cross-scale patch-level self-supervision for training purposes.

% our SS-PRL exploits cross-scale correlation loss $L_{cross}$ to integrate the extracted information at each scale and explore semantic dependencies across multiple scales.

% We utilize the cluster predictions $P_s$ at every local patch scale $s$ obtained from \eqref{eqn:probability} to predict the cluster assignments of the augmented global image level (\ie, $Q_0$ and $Q_0^{\prime}$).
% , which are denoted by $Q_0 = [q_{0,1}]$ and $Q_0^{\prime} = [q_{0,1}^{\prime}]$.
% By exploiting this objective function, the feature extractor $f_\theta$ is guided to extract 
% \textcolor{blue}{coherent}
% information across aggregated regional representations at different scales.
% Moreover, the semantic label structure will be explored with the multi-level prototypes since this 
% \textcolor{blue}{correlation}
% objective naturally considers the relations/dependencies between clusters from different levels. 
More specifically, we perform average pooling across all $M_s$ representation vectors in $P_s$ and $P_s^{\prime}$ from level $s$, resulting in $\mu({P}_s)$ and $\mu({P^\prime_s})$, respectively. We then apply a set of cross-scale correlation learners $g_\phi = \{g_{\phi,s}\}_{s=1}^{S}$, one for each scale, to project $\mu({P}_s)$ and $\mu({P^\prime_s})$ onto the representation space of $p_{0}$ and $p_{0}^{\prime}$, i.e., at the global image level. As a result, our cross-scale correlation loss $L_{cross}$ can be formulated as:
\begin{equation}
\label{Eq:Cross-Level Loss}
\begin{split}
L_{cross} = \sum_{s=1}^{S}\beta_s(\mathit{CE}(Q_0, g_{\phi,s}(\mu({P}_s))) + \\
\mathit{CE}(Q_{0}^{\prime}, g_{\phi,s}(\mu({P^\prime_s})))
% L_{cross} &=- \sum_{s=1}^{S}\beta_s({q_{0}}^{\top}\log{g_{\theta,s}({\overline{p_{s}})}} +  
% {q_{0}^{\prime}}^{\top}\log{g_{\theta,s}({\overline{p_{s}}^{\prime})}})
,
\end{split}
\end{equation}

\noindent where $\mathit{CE}$ is the cross-entropy loss, and $\beta_s$ balances cross-scale correlation losses across different scales.

It is worth noting that, learning pyramid representations for different patch-level scale pairs not only encourages the feature extractor $f_\theta$ to exploit the patch-level information in an image, it also aggregates the fine-grained semantics for matching the global ones (via $g_\phi$) presented in an image. As confirmed in our experiments, this self-supervised learning strategy allows us to fine-tune $f_\theta$ for downstream tasks associated with multi-labeled images.

\begin{table*}[t]
\begin{center}
\scalebox{0.73}{
\begin{tabularx}{1.33\textwidth}{l@{\extracolsep{\fill}}*{13}{c}}
\toprule
 & & \multicolumn{12}{c}{Mask R-CNN R50-FPN COCO 15k} \\
\cmidrule(l){3-14}
 & & \multicolumn{6}{c}{Pretrained on COCO} & \multicolumn{6}{c}{Pretrained on ImageNet} \\
\cmidrule(lr){3-8} \cmidrule(l){9-14}
Method & & $\text{AP}^{bb}$ & $\text{AP}^{bb}_{50}$ & $\text{AP}^{bb}_{75}$ & $\text{AP}^{mk}$ & $\text{AP}^{mk}_{50}$ & $\text{AP}^{mk}_{75}$ & $\text{AP}^{bb}$ & $\text{AP}^{bb}_{50}$ & $\text{AP}^{bb}_{75}$ & $\text{AP}^{mk}$ & $\text{AP}^{mk}_{50}$ & $\text{AP}^{mk}_{75}$ \\
\midrule
Random Init. & & 11.5 & 21.3 & 11.3 & 10.8 & 19.7 & 10.7 & 11.5 & 21.3 & 11.3 & 10.8 & 19.7 & 10.7 \\
\midrule
MoCo v2 \cite{he2020momentum} & \multirow{3}{*}{\shortstack{\textit{general-purpose}\\\textit{SSL}}} & 17.0 & 30.6 & 17.2 & 15.9 & 28.4 & 15.9 & 21.1 & 36.7 & 21.8 & 19.9 & 34.4 & 20.4 \\
SwAV \cite{caron2020unsupervised} & & 18.1 & 33.7 & 17.6 & 17.3 & 31.5 & 17.1 & 23.1 & \underline{41.2} & 23.4 & 22.1 & \underline{38.6} & 22.5 \\
BYOL \cite{grill2020bootstrap} & & 17.4 & 31.5 & 17.4 & 16.2 & 29.2 & 16.1 & 21.4 & 37.5 & 22.1 & 20.1 & 35.1 & 20.6 \\
\midrule
DenseCL \cite{wang2021dense} & \multirow{4}{*}{\shortstack{\textit{dense prediction}\\\textit{SSL}}} & \textbf{20.2} & \underline{35.4} & \textbf{20.8} & \underline{18.9} & \underline{33.0} & \textbf{19.3} & 21.9 & 38.0 & 22.9 & 20.7 & 35.8 & 21.3 \\
DetCo \cite{xie2021detco} & & 15.6 & 29.7 & 14.8 & 14.8 & 27.3 & 14.4 & 20.9 & 38.1 & 20.9 & 19.9 & 35.3 & 19.9 \\
MaskCo \cite{zhao2021self} & & \underline{18.5} & 32.9 & 18.7 & 17.3 & 30.7 & 17.4 & 20.6 & 35.6 & 21.5 & 19.5 & 33.4 & 20.0 \\
InsLoc \cite{yang2021instance} & & 17.5 & 31.5 & 17.6 & 16.5 & 29.3 & 16.6 & \underline{23.5} & 40.5 & \textbf{24.7} & \underline{22.2} & 38.1 & \underline{22.9} \\
\midrule
\textbf{SS-PRL (ours)} & & \textbf{20.2} & \textbf{36.7} & \underline{20.2} & \textbf{19.1} & \textbf{34.3} & \underline{19.0} & \textbf{23.6} & \textbf{42.5} & \underline{24.0} & \textbf{22.7} & \textbf{39.7} & \textbf{23.1} \\
\bottomrule
\end{tabularx}}
\end{center}
\vspace{-2mm}
\caption{\textbf{Downstream object detection and instance segmentation tasks on COCO.} We report the bounding box AP ($\text{AP}^{bb}$) for object detection and the mask AP ($\text{AP}^{mk}$) for instance segmentation on COCO. All methods are pre-trained on COCO for 200 epochs or ImageNet for 100 epochs and then fine-tuned for the above tasks on COCO for 15k iterations. Note that \textit{Random Init.} denotes the detector trained from scratch (\ie the encoder is randomly initialized without any pre-training). The best results in each category are in \textbf{bold}, and the second-best ones are \underline{underlined}.}
\vspace{-3mm}
\label{table:inst_seg_15k}
\end{table*}

\begin{table}[t]
\begin{center}
% \vspace{-6mm}
\scalebox{0.75}{
\begin{tabularx}{0.49\textwidth}{lY}
\toprule
Prototype & mAP \\
\midrule
Baseline & 79.2 \\
\midrule
Shared across all scales & 79.4 \\
Learned \& correlated across scales & \textbf{80.5} \\
\bottomrule
\end{tabularx}}
\quad
\scalebox{0.75}{
\begin{tabularx}{0.49\textwidth}{lY}
\toprule
Method & mAP \\
\midrule
Baseline & 79.2 \\
\midrule
SS-PRL w/ $L_{pyr}$ only & 79.5 \\
SS-PRL w/ $L_{cross}$ only & 79.8 \\
\midrule
Full SS-PRL ($L_{pyr}$ + $L_{cross}$) & \textbf{80.5} \\
\bottomrule
\end{tabularx}}
\end{center}

\caption{
\textbf{Ablation Studies on the derived patch-level prototypes (top) and the proposed loss functions (down).}
Note that \emph{Shared across all scales} indicates the same prototypes learned across patch scales (\ie, same $C_s$ at different patch scales in Fig.~\ref{fig:framework}). We see
that prototypes learned from each scale and enforced by our cross-scale correlation would be desirable. And, SS-PRL achieves the best results\ when both $L_{pyr}$ and  $L_{cross}$ are introduced.}
\vspace{-2mm}
\label{table:ablation}
\end{table}

\subsection{Pre-Training and Fine-Tuning Stages} \label{ssec:summary}
\noindent \textbf{Self-supervised pre-training of $f_\theta$ and $g_\phi$.}
Overall, the full objective function $L$ for pre-training feature extractor $f_\theta$ and the cross-scale correlation learner $g_\phi$ can be summarized below:
\begin{equation}
\label{Eq:Overall Loss}
L= L_{pyr} + \lambda L_{cross},
\end{equation}
where $\lambda$ acts as the weight to balance the two terms, and is set as 1.0 throughout our work. On the other hand, we select the same values for $\alpha_s$ and $\beta_s$ in \eqref{eqn:L_pyr} and \eqref{Eq:Cross-Level Loss} for simplicity (we set these hyperparameters as 1.0 for $s=0$ and 0.25 for other scales to balance the influence of different levels). The effectiveness of each loss is later confirmed by the ablation study in Section~\ref{Ablation:Proposed loss}, and the pseudo-code is summarized in the supplementary material. \\
% Algorithm~\ref{alg:Algorithm1}.

\noindent \textbf{Supervised fine-tuning for $f_\theta$.}
Once the feature extractor $f_\theta$ is pre-trained via our proposed SS-PRL, we then fine-tune it to downstream tasks associated with multi-label images in a supervised fashion. For example, as presented in Section~\ref{sec:experiments}, we adapt the pre-trained $f_\theta$ (\eg, with the architecture as ResNet-50~\cite{he2016deep}) to multi-label image classification, object detection, and segmentation tasks using different amounts of images with ground truth annotation. Please see the next section for the thorough experiments on these tasks and comparisons to state-of-the-art SSL methods.

%% file: 4_experiment.tex
\section{Experiments}
\label{sec:experiments}

%------------------------------------------------------------------------

\subsection{Datasets and Experimental Setups}

\paragraph{Pre-training Dataset.}
% \noindent \textbf{Pre-training datasets.}
We consider MSCOCO~\cite{lin2014microsoft} and ImageNet~\cite{deng2009imagenet}. For MSCOCO~\cite{lin2014microsoft}, COCO~\texttt{train2014}~\cite{lin2014microsoft}, which contains $\sim$83k images, is used for SSL pre-training, and we train all methods for 200 epochs with a batch size of 128. 
As for ImageNet~\cite{deng2009imagenet}, we utilize the training set with $\sim$1.28M training images for SSL pre-training, and train  methods for 100 epochs with a batch size of 256. Our image pyramids contain three patch scales (\ie, $s=0,1,2$) in all experiments. The patch sets consist of 4 patches ($M_1 = 4$) at scale $s = 1$ and 9 patches ($M_2 = 9$) at scale $s = 2$. Further training details such as data generation and hyperparameter selection are provided in our supplementary material. 
% -----------------------------
\label{ssec:setting}

\begin{figure*}[t]
\begin{center}
\includegraphics[width=0.80\linewidth]{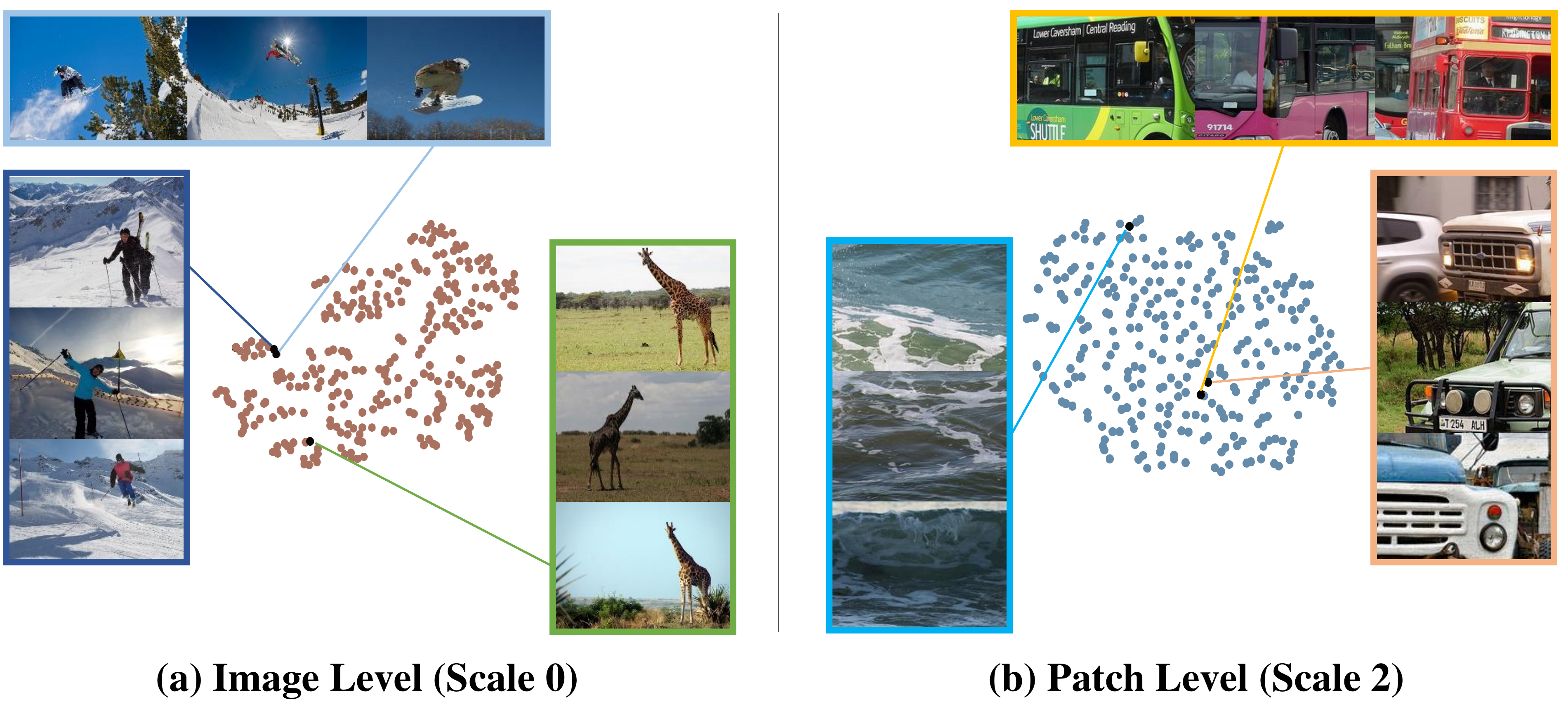}
\end{center}
\vspace{-3mm}
\caption{\textbf{t-SNE visualization of the learned prototypes on COCO.} We visualize the learned prototype at the corresponding scale, with selected images associated with each prototype illustrated. (a) At scale $s = 0$, nearby prototypes show similar semantic meanings of scenes (\eg snowfield). (b) At scale $s = 2$, nearby prototypes are semantically related object-level information (\eg, cars).} %Moreover, prototypes far from each other are seen to describe dissimilar visual concepts.}
\label{fig:tsne}
\vspace{-2mm}
\end{figure*}

\noindent \textbf{Evaluation protocol.}
We evaluate the pre-trained models by fine-tuning on downstream multi-label classification, object detection, and segmentation tasks using MSCOCO~\texttt{train2014}~\cite{lin2014microsoft} and PASCAL VOC~\cite{everingham2010pascal}. For multi-label classification task, we follow the linear evaluation setting~\cite{misra2020self} to train a linear multi-label classifier on top of the fixed pre-trained backbone network (\eg, Resnet-50) on COCO \texttt{train2014}~\cite{lin2014microsoft} and VOC \texttt{trainval07}~\cite{everingham2010pascal}, and then report mean average precision (mAP) on COCO \texttt{val2014}~\cite{lin2014microsoft} and VOC \texttt{test2007}~\cite{everingham2010pascal}. We also follow the semi-supervised setting and randomly sample 1\%, 10\%, and 100\% labeled data from COCO \texttt{train2014}~\cite{lin2014microsoft} (which is $\sim$0.8k, $\sim$8k, and $\sim$83k images) to fine-tune the whole network for 20 epochs, and then report mAP on COCO \texttt{val2014}~\cite{lin2014microsoft}.

As for object detection and instance segmentation tasks, we pre-train and fine-tune a Mask R-CNN~\cite{he2017mask} detector with FPN~\cite{lin2017feature} backbone on COCO \texttt{train2014} \cite{lin2014microsoft} and evaluating on COCO \texttt{val2014}~\cite{lin2014microsoft}. Note that, synchronized batch normalization is applied in the backbone network, FPN, and prediction heads during training. We report the results of detectors with $15k$ training iterations to compare the transfer ability of each SSL pre-training method.
Due to page limitations, we provide quantitative comparisons on the downstream semantic segmentation task in our supplementary material. 

% To evaluate the learned representations, we follow the linear evaluation protocol~\cite{misra2020self}, in which a linear multi-label classifier is trained with the fixed pre-trained feature extractor, and mean average precision (mAP) is used to evaluate the representation quality. We use two datasets, COCO \cite{lin2014microsoft} and VOC \cite{everingham2010pascal}, to evaluate the pre-trained models. For COCO, we train the linear classifier on the same \texttt{train2014} used for pre-training, and then report mAP on COCO \texttt{val2014}, which contains $\sim$41k images. In addition, we choose VOC to evaluate the transfer learning performance of our pre-training approach. We train the linear classifier on VOC \texttt{trainval07}, which
% contains $\sim$5k images, and then report mAP on the other $\sim$5k images in VOC \texttt{test2007} set.
% % segmentation in supplement %
% Due to page limitation, we provide quantitative comparisons on downstream \emph{segmentation} tasks in our supplementary material.

%------------------------------------------------------------------------

\subsection{Quantitative Evaluation}

% \begin{table}[t]
% \begin{center}
% \scalebox{0.9}{
% \begin{tabularx}{0.65\linewidth}{l *2Y}
% \toprule
%  & \multicolumn{2}{c}{Semantic Segmentation (mIoU)} \\ 
% \cmidrule(l){2-3}
% Pre-training Method & COCO $\rightarrow$ VOC & IN $\rightarrow$ VOC \\
% \midrule
% Random Init. & 40.7 & 40.7 \\
% \midrule
% MoCo v2 \cite{he2020momentum} & 57.3 & 65.7 \\
% SwAV \cite{caron2020unsupervised} & 56.1 & 60.7 \\
% BYOL \cite{grill2020bootstrap} & 54.1 & 61.8 \\
% \midrule
% DenseCL \cite{wang2021dense} & \underline{63.2} & 69.0 \\
% DetCo \cite{xie2021detco} & 43.2 & \\
% MaskCo \cite{zhao2021self} & \underline{59.8} & 65.7 \\
% InsLoc \cite{yang2021instance} & 56.1 & 67.1 \\
% \midrule
% \textbf{SS-PRL (ours)} & \underline{62.4} & 60.4 \\
% \bottomrule
% \end{tabularx}}
% \end{center}
% \caption{\textbf{Downstream semantic segmentation task on the PASCAL VOC dataset.} Note that we report the mean IoU (mIoU) on VOC with fine-tuned FCN models. All methods are pre-trained on COCO with 200 epochs. Top-3
% best pre-training methods are \underline{underlined}.}
% \label{table:Semantic_Seg}
% \end{table}

\paragraph{Multi-label classification with fine-tuned linear classifiers.}
% \subsubsection{Multi-Label Classification in Linear Evaluation Setting.}

We first perform downstream multi-label image classification with fine-tuned linear classifiers and compare our results with existing general-purpose~\cite{he2020momentum,caron2020unsupervised,grill2020bootstrap} and dense prediction based~\cite{wang2021dense,xie2021detco,zhao2021self,yang2021instance} self-supervised learning methods on two commonly-used public benchmarks, COCO~\texttt{train2014}~\cite{lin2014microsoft} and VOC~\cite{everingham2010pascal}. In Table~\ref{table:main}, we observe that our SS-PRL outperforms state-of-the-art SSL approaches on multi-label classification benchmarks when pre-trained on COCO dataset~\cite{lin2014microsoft}. Moreover, SS-PRL surpasses all SSL methods by a significant margin when pre-trained on ImageNet~\cite{deng2009imagenet} by obtaining 63.8\% and 85.4\% mAP on COCO~\cite{lin2014microsoft} and VOC~\cite{everingham2010pascal}, respectively.
With the proposed pyramid representation learning, we are able to obtain better results than previous SSL methods~\cite{he2020momentum,caron2020unsupervised,grill2020bootstrap} that are not designed to handle patch or object-level information. It can be seen that our method also outperforms SSL methods that integrate local information for exploiting data discrimination~\cite{wang2021dense,xie2021detco,zhao2021self,yang2021instance} with large margins. \\

\noindent \textbf{Multi-label classification in semi-supervised settings.}
Table~\ref{table:semi} compares SS-PRL results with previous SSL methods in the semi-supervised settings of multi-label classification by sampling 1\% and 10\% labeled data. SS-PRL significantly improves over the state-of-the-art in most settings, showing the prowess when transferred to datasets with limited annotation. We also provide results when fine-tuned with 100\% labeled data, where we outperform the randomly initialized model by 20.4\% and 24.9\% mAP. From this experiment, the effectiveness of our model for multi-label image classification can be successfully confirmed. \\

\noindent \textbf{Object detection and instance segmentation.}
The results of object detection and instance segmentation tasks on COCO~\cite{lin2014microsoft} with $15k$ training iterations are reported in Table~\ref{table:inst_seg_15k}. SS-PRL outperforms existing general-purpose SSL methods and achieves comparable or even better results with dense prediction based SSL methods when pre-trained on both COCO~\cite{lin2014microsoft} and ImageNet~\cite{deng2009imagenet}. The above results exhibit the impressive ability of SS-PRL for downstream dense prediction tasks at object or instance levels.
% We first perform downstream multi-label image classification tasks and compare our results with existing general-purpose~\cite{he2020momentum,caron2020unsupervised,grill2020bootstrap} and dense prediction based~\cite{wang2021dense,xie2021detco} self-supervised learning methods on two commonly-used public benchmarks, COCO~\texttt{train2014} and VOC~\cite{everingham2010pascal}, as shown in Table~\ref{table:main}. For fair comparisons, all models are pre-trained on COCO~\texttt{train2014} for 200 epochs with a batch size of 128. For our quantitative comparisons, we construct pyramid views consisting of 3 different spatial scales (\ie, $S = 2$). In Table~\ref{table:main}, we observe that our SS-PRL outperforms state-of-the-art SSL approaches on multi-label classification benchmarks and achieves 61.1\% and 80.1\% mAP on COCO and VOC, respectively. We also narrow the performance gap with fully supervised pre-training ones (as upper bound). With the proposed pyramid representation learning, we are able to obtain better results than previous SSL methods~\cite{he2020momentum,caron2020unsupervised,grill2020bootstrap} that do not address the presence of multiple objects in their design. Our method also outperforms SSL methods that integrate local information into instance discrimination scheme~\cite{wang2021dense,xie2021detco} with a large margin.

%------------------------------------------------------------------------

\subsection{Ablation Study}
\label{ssec:ablation}

We now conduct ablation studies and parameter analysis to better understand how each component of SS-PRL contributes to the overall performance in downstream multi-label classification tasks. We pre-train models on the COCO~\cite{lin2014microsoft} dataset and report the mAP on VOC~\cite{everingham2010pascal} for evaluation. We adopt SS-PRL trained with global images only (\ie, $s = 0$) as our baseline. \\
% In this subsection, we conduct an ablation study and parameter analysis to better understand how each component of SS-PRL contributes to the overall performance in the downstream multi-label image classification task on VOC dataset. We report the mAP for evaluation, and adopt SS-PRL only trained with global augmented views (\ie, $s=0$) as our baseline. While we follow the setups introduced in Section~\ref{ssec:setting} for training network models, only \textit{two} image scales (\ie, $s = 0, 1$) are considered. For the completeness of analysis, we have results using three image scales in the supplementary material.

%------------------------------------------------------------------------

%\subsubsection{Ablation Study}

\begin{figure*}[t]
\begin{center}
\includegraphics[width=0.8\linewidth]{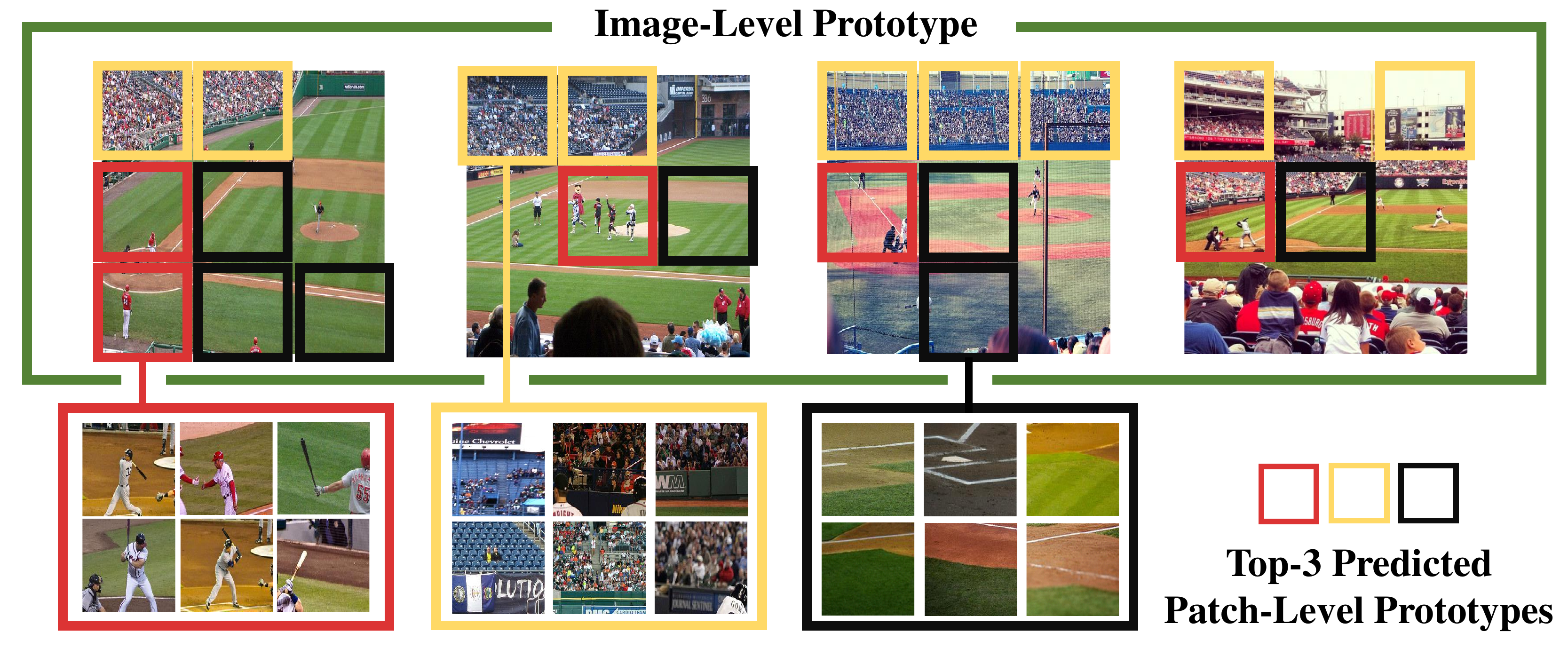}
\end{center}
\vspace{-3.5mm}
\caption{\textbf{Correlation between prototypes across different levels.} We randomly choose an image-level (scale 0) prototype from COCO (marked in \textcolor{OliveGreen}{green}) and visualize its top-3 corresponding patch-level prototype predictions at scale 2 (marked in \textcolor{BrickRed}{red}, \textcolor{Dandelion}{yellow} and \textcolor{black}{black}). With examples of the three selected patch-level prototypes shown at the bottom row, we observe that the patch-level prototypes distinctively represent fine-grained visual concepts which are related to those of the image-level prototypes.}
\label{fig:cluster}
\vspace{-2.2mm}
\end{figure*}

\noindent \textbf{Learning of patch-level prototypes.}
The patch-level prototypes $C_s$ introduced in Section~\ref{sssec:pyramid} provide semantic cues of inherited label dependencies observed in training data, and ensure the feature extractor $f_\theta$ to exploit meaningful regional information at each patch scale from an image. In Table~\ref{table:ablation}, we report the linear evaluation results of SS-PRL trained with prototypes $C_s$ learned within and across scales $s$. It can be seen that mAP drops by 1.1\% when prototypes are shared across different scales. This indicates that the prototypes at different patch scales capture the hierarchical semantic/label dependencies of the dataset that is crucial to the downstream tasks with multi-labeled data. Additional visualization for such learned prototype sets will be shown in Figure~\ref{fig:tsne}, \ref{fig:cluster} and discussed in Section~\ref{ssec:visualization}. 

%------------------------------------------------------------------------
\noindent \textbf{\\Loss functions.}
\label{Ablation:Proposed loss}
To analyze the effectiveness of each developed loss function (\ie, the pyramid representation learning loss $L_{pyr}$ and the cross-scale correlation loss $L_{cross}$), we conduct an ablation study on the VOC dataset~\cite{everingham2010pascal}. Table~\ref{table:ablation} reports the performance of multi-label image classification tasks using linear evaluation protocol. 
We observe limited performance improvement (+0.3\%) when the model is only trained with the pyramid loss $L_{pyr}$. This is due to the fact that the objectives enforced at each scale are not guaranteed to be mutually related, restricting the discriminative capability.
When cross-scale correlation loss $L_{cross}$ is included, we observe the performance boosts up 0.6\% mAP compared to the baseline. This indicates the importance of exploring the correspondence of pyramid features across each level to derive discriminative yet coherent representations. The best results (+1.3\%) are obtained by our full SS-PRL which considers both $L_{pyr}$ and $L_{cross}$, exploiting semantic concepts and correspondence within and across patch scales.
\subsection{Visualization}
\label{ssec:visualization}

%\subsubsection{Clusters learned at different levels.}

% Figure~\ref{fig:prototype} visualizes the semantic concepts portrayed by our learned prototypes at different patch levels on the COCO dataset~\cite{lin2014microsoft}. We observe that 
% %image-level (scale 0) prototypes tend to capture the concept of scenes that include multiple objects or items (\eg motorcycle with a rider or steam train blowing smoke), while each region-level prototype (scale 2) tends to capture a single object in an image (\eg a tire or rail). 
% prototypes at scale 0 tend to encapsulate the concept of entire scenes that include multiple objects or items (\eg motorcycle with a rider or steam train blowing smoke), while prototypes at scale 1 tend to capture a single object in an image (\eg a motorcycle or a train), and prototypes at scale 2 tend to portray a fine-grained item (\eg a tire or rail).
% It shows that SS-PRL is able to discover the hierarchical semantic structure of the dataset with our proposed pyramid representation learning. 
% ---------------------figure------------------

% -------------------paragraph-------------------
%\subsubsection{Dependency of prototypes at different patch levels.}
\noindent \textbf{Prototypes learned at each scale.}
To further visualize and relate the prototypes learned at different scales, we visualize the learned global image-level and local patch-level prototypes using t-SNE \cite{van2008visualizing} and show example results in Figures~\ref{fig:tsne} (a) and (b), respectively. In both cases, nearby prototypes show semantically related visual concepts compared to prototypes far apart. At the image level (a), nearby prototypes share similar semantics of \emph{scenes} (\eg skiing and snowboarding). At the patch level (b), prototypes close to each other show related semantic concepts of \emph{objects} (\eg two different parts of a car). On the contrary, two prototypes far apart represent different semantic meanings at both levels (\eg snowfield vs. grassland and cars vs. ocean). This demonstrates that our method would be able to discover semantic dependencies at different patch scales. 

% ----------------------------------------------
\noindent \textbf{\\Dependency of prototypes across different levels.}
Finally, we visualize the correlation dependency between an image-level prototype and the associated patch-level prototypes in Figure~\ref{fig:cluster}. Specifically, we divide all the images from a randomly chosen image-level prototype into patches and generate their corresponding patch-level prototype predictions by SS-PRL. All the results of predictions are counted and the top-3 patch-level prototypes that are most frequently predicted will be visualized. The patch-level prototypes correspond to three different iconic elements (\ie,  fields, audience, and players) that further mine the fine-grained semantics from the image-level prototype (\ie, baseball games), showing the semantic correspondence of predictions across image and patch levels.

%------------------------------------------------------------------------

%% file: 5_conclusion.tex
\section{Conclusion}

In this paper, we presented Self-Supervised Pyramid Representation Learning (SS-PRL) for pre-training deep neural networks, with the goal to facilitate downstream vision tasks at object, instance, or pixel levels. By deriving pyramid representations and learning prototypes at each patch level, our SS-PRL is able to exploit the inherent semantic information within and across image scales via self-supervision. This is achieved by our introduced cross-scale patch-level correlation learning, which aggregates and associates the knowledge across different scales, observing and enforcing the dependency between pyramid representations across patch levels. We conduct a wide range of experiments, including the tasks of multi-label image classification, object detection, and instance segmentation, which support the use of our SS-PRL as a desirable pre-training strategy. With visualization of the learned representations and ablation studies, the design of the proposed SS-PRL can be properly verified.